\documentclass{article}
\usepackage{ijcai21}

\usepackage{times}
\usepackage{soul}
\usepackage{url}
\usepackage[hidelinks]{hyperref}
\usepackage[utf8]{inputenc}
\usepackage[small]{caption}
\usepackage{graphicx}
\usepackage{amsmath}
\usepackage{amsthm}
\usepackage{booktabs}
\usepackage{algorithm}
\usepackage{algorithmicx}
\usepackage{algpseudocode}
\usepackage{amssymb}
\newtheorem{theorem}{Theorem}
\newtheorem{definition}{Definition}

\title{Optimize Neural Fictitious Self-Play in Regret Minimization Thinking}

\author{
Yuxuan Chen$^1$
\and
Li Zhang$^2$\and
Shijian Li$^2$\And
Gang Pan$^4$
\affiliations
$^1$College of Computer Science, Zhejiang University\\
$^2$Second Affiliation\\
$^3$Third Affiliation\\
$^4$Fourth Affiliation
\emails
\{yuxuan\_chen, zhangli85, shijianli, gpan, zhijie\_pan, chenxli.cs\}@zju.edu.cn,
}

\begin{document}

\maketitle

\begin{abstract}
  Optimization of deep learning algorithms to approach Nash Equilibrium remains a significant problem in imperfect information games, e.g. StarCraft and poker. Neural Fictitious Self-Play (NFSP) has provided an effective way to learn approximate Nash Equilibrium without prior domain knowledge in imperfect information games. However, optimality gap was left as an optimization problem of NFSP and by solving the problem, the performance of NFSP could be improved. In this study, focusing on the optimality gap of NFSP, we have proposed a new method replacing NFSP's best response computation with regret matching method. The new algorithm can make the optimality gap converge to zero as it iterates, thus converge faster than original NFSP. We have conduct experiments on three typical environments of perfect-information games and imperfect information games in OpenSpiel and all showed that our new algorithm performances better than original NFSP.
\end{abstract}

\section{Introduction}
With the rapid development of deep reinforcement learning, AI algorithms have already beaten human experts in most perfect-information games like Go~\cite{DBLP:journals/nature/SilverHMGSDSAPL16,DBLP:SilverSSAHGHBLB17}. But still there remains challenges to solve imperfect-information dynamic games. In the recent years, many efforts have been made, professional players were beaten by AI in StarCraft2~\cite{DBLP:conf/gecco/ArulkumaranCT19} and Dota (OpenAI-Five)~\cite{DBLP:journals/corr/abs-1912-06680}. Though such researches exhibit remarkable performances, there are few works to learn an optimal policy from a view of game theory. Game theory is essential for proving and guaranteeing the quality of learned policies.

Game theory~\cite{10.2307/1969529} is a study about real world competitions. It studies how to maximize the benefits with different and dynamic policies. Nowadays, game theory plays a more and more important role in designing algorithms to solve real-world competition problems such as transaction and traffic control~\cite{DBLP:journals/aim/Sandholm10,DBLP:journals/jair/BosanskyKLP14}.

In game theory, Nash Equilibrium~\cite{10.2307/1969529} would be a choice of optimal policy profile in a game. Nash Equilibrium is a state where all players select their optimal strategies and no one can gain extra profit by changing their policies. Finding and reaching Nash Equilibrium in imperfect information games need particular algorithm designs. There are two commonly used methods to reach Nash Equilibrium: game tree search method and fictitious play (FP) method. Tree search methods such as counterfactual regret minimization (CFR)~\cite{Bowling145} and Monte Carlo Tree Search (MCTS)~\cite{DBLP:journals/tciaig/BrownePWLCRTPSC12} traverse the game tree and calculate the value of each nodes. Tree search methods always cost a lot of time to sample enough data and require large space to ensure the recursive tree search process. In a different way, fictitious play method improves its knowledge while agents play against with each other, which requires less space.

In fictitious play(FP)~\cite{brown1951iterative}, a player iteratively improves its best response strategy by fighting against its opponents in timesteps of simulated games. Then the average strategy of its history best responses will converge to Nash Equilibrium after enough iterations. Heinrich et al.~\cite{FSP} extended the FP algorithm to solving extensive-form games and then introduced a sampling based machine learning approach called Fictitious Self-Play (FSP) to make training easier. FSP updates the average strategy with sampling based supervised learning methods and approximates the best response with reinforcement learning methods. Reinforcement learning have been demonstrated to achieve excellent results on a range of complex tasks including games~\cite{DBLP:conf/icml/OhCSL16} and continuous control~\cite{DBLP:conf/icml/SchulmanLAJM15,DBLP:journals/jmlr/LevineFDA16}.

Heinrich and Silver~\cite{NFSP} developed FSP with deep neural network. Neural Fictitious Self-Play, which introduce Deep Q-network (DQN)~\cite{DBLP:journals/nature/MnihKSRVBGRFOPB15,DBLP:journals/corr/abs-1806-06953} to approximate its best response and supervised learning network to update its average strategy. In NFSP, player produce best responses according to a mixture of opponents' average policies and best responses. NFSP is the first end-to-end reinforcement learning method that learns approximate Nash Equilibrium in imperfect information games without any prior knowledge like ELF~\cite{DBLP:conf/nips/TianGSWZ17,DBLP:journals/corr/abs-1902-02004}.

Because NFSP update strategies according to opponents' average strategy iteratively, there exists optimality gap which best response (reinforcement learning) needs to improve at every update. In this study, we find and prove that the optimality gap could be transferred into a monotonical form by applying regret matching~\cite{Sergiu2000A} methods. In practice, we use Advantage-based Regret Minimization (ARM)~\cite{ARM} as the regret matching method replacing the computation of NFSP's best response
We have conduct experiments in 3 different environments provided by OpenSpiel~\cite{OpenSpiel} and all experiments show that the new algorithm outperforms NFSP and converges faster than NFSP.

\section{Preliminaries}

In our work, intending to reduce optimality gap, we apply regret matching method to NFSP's best response computation and make best response optimized at every update iteration. Optimized best response would improve behavioural strategy and then make it better of approaching nash equilibrium.

Therefore, in this section we will briefly introduce some backgrounds of our method and strategy update process of NFSP. At last, we will point out the focus of our paper, optimality gap.

\subsection{Backgrounds}

This subsection provides some background knowledge of our method, including extensive-form game and best response.

Extensive-form game is a model of sequential interaction involving multiple agents. Each player's goal is to maximize his payoff in the game. A player's \textbf{behavioural strategy} $\pi^i$ determines a probability distribution over actions given an information state. Each player choose a behavioural strategy while playing games. We assume games with perfect recall, i.e. each player's current information state $s^i_t$ implies knowledge of the sequence of his information states and actions, $s^i_1,a^i_1,s^i_2,a^i_2,\ldots,s^i_t$, that led to this information state.
 A \textbf{strategy profile} $\pi=(\pi^1,\ldots,\pi^n)$ is a collection of strategies for all players. $\pi^{-i}$ refers to all strategies in $\pi$ except $\pi^i$. Based on the game's payoff function $R$, $R^i(\pi)$ is the expected payoff of player $i$ if all players follow the strategy profile $\pi$. 

The set of \textbf{best response} of player $i$ to their opponents' strategies $\pi^{-i}$ is $b^i(\pi^{-i})= \arg\max_{\pi^i}(R^i(\pi^i, \pi^{-i}))$. The set of $\epsilon$-best response to the strategy profile $\pi^{-i}$ is defined as $b^i_\epsilon(\pi^{-i}) = \{\pi^i:R^i(\pi^i, \pi^{-i})\geq R^i(b^i(\pi^{-i}), \pi^{-i})-\epsilon\}$ for $\epsilon > 0$. A Nash equilibrium of an extensive-form game is a strategy profile $\pi$ such that $\pi^i\in b^i(\pi^{-i})$ for all player $i$. An \textbf{$\epsilon$-Nash equilibrium} is a strategy profile $\pi$ that $\pi^i\in b^i_\epsilon(\pi^{-i})$ for all player $i$.

\subsection{Neural Fictitious Self-Play (NFSP)}

Neural Fictitious Self-Play (NFSP) is an effective end-to-end reinforcement learning method that learns approximate Nash Equilibrium in imperfect information games without any prior knowledge. NFSP's strategy consists of two strategies: best response and average strategy. Average strategy averages past responses and is the exact strategy to approach Nash Equilibrium, equivalent to behavioural strategy. NFSP iteratively updates best response and average strategy to approach approximate Nash Equilibrium. Our method replaces best response computation in Neural Fictitious Self-Play (NFSP) with regret matching, based on the strategy update rule of NFSP. In this subsection, we will introduce NFSP and provide details of the strategy update process of NFSP. For a more detailed exposition we refer the reader to ~\cite{FSP,NFSP}.

NFSP extends fictitious self-play (FSP) with deep neural networks based on fictitious play (FP) theorem.
In NFSP, fictitious players choose best response to their opponents' average behaviour. 
NFSP approximates best response in extensive-form games, it replaces the best response computation and the average strategy updates with reinforcement learning and supervised learning respectively. NFSP agent's strategy update process at iteration $t+1$ follows the rule of generalized weakened fictitious play~\cite{DBLP:journals/geb/LeslieC06} which is guaranteed to approach approximate Nash Equilibrium:

$\Pi^i_{t+1}\in(1-\alpha_{t+1})\Pi_t^i+\alpha_{t+1}B^i_{\epsilon_t}(\Pi_t^{-i})$

where $B^i_\epsilon$ is the $\epsilon$-best response of player $i$, $\Pi^i$ is the average strategy of player $i$, and $\alpha_{t+1}<1$ is a mix probability of best response and average strategy.

\begin{figure}[h]
  \centering
  \includegraphics[width=\linewidth]{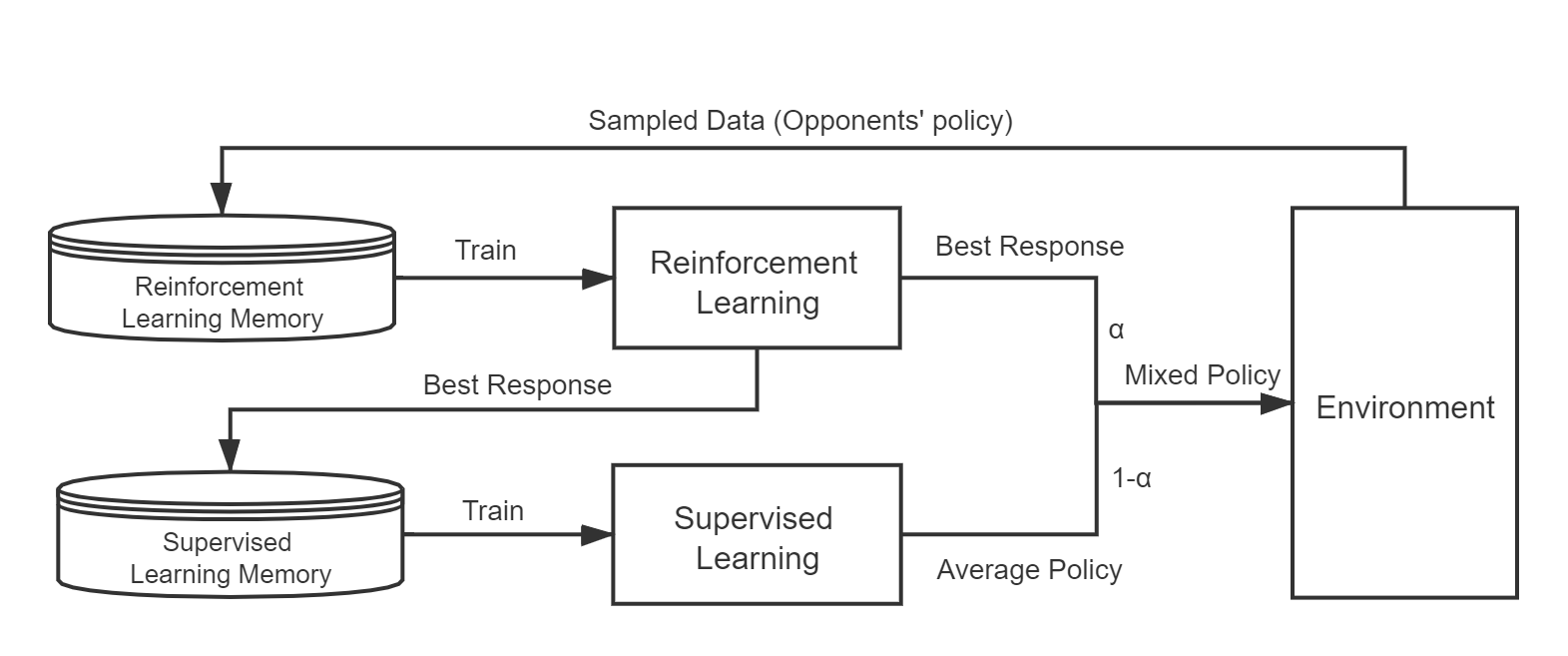}
  \caption{the overall structure of NFSP (one agent)}
  \label{fig:NFSP}
\end{figure}

As figure shows, NFSP maintains two strategy networks, $B_{t+1}$ to learn the best response to the average policy $\Pi_t$, and $\Pi_{t+1}$ to learn the average policy of the best response $B_{t+1}$ and $\Pi_t$ in a determined proportion.

The resulting NFSP algorithm is as follows. Each agent $i$ has its reinforcement learning network $B^i$ as its best response, supervised learning network $\Pi^i$ as its average strategy, and supervised learning replay buffer 
$\mathcal{M}^i_{SL}$. At the very start of a game, each agent decides whether it chooses $B^i$ or $\Pi^i$ as its current strategy in this episode, with probability $\alpha$ and $1-\alpha$ respectively. At each timestep in this episode, agents sample an action from the selected strategy and if the selected startegy is best response, a tuple of observation and action taken is stored in $\mathcal{M}^i_{SL}$. $B^i$ is trained as it maximizes the expected cumulative reward against $\Pi^{-i}$ and $\Pi^i$ is trained as it demonstrates the average strategy over the past best responses. The two networks update after a set number of episodes if the replay buffer size is bigger than a certain number.

Though NFSP is an efficient end-to-end algorithm, there still exists a problem of optimality gap that makes convergence not fast as expected.

Optimality gap $\epsilon$ refers to $\epsilon$-best response at every update. It is expected that optimality gap $\epsilon$ monotonically reduces as iteration goes, but $\epsilon$ is actually asymptotically decaying according to \cite{FSP}. In the result, optimality gap is not reducing in a monotonical way. In section 3, we explain why optimaligy gap reduces asymptotically and propose a method to make it reduce monotonically, faster than original method.
\section{Optimization}
In this section we will discuss details of optimality gap under NFSP's stratrgy update rule, introduce the method to make optimality gap reduce monotonically and provide theoretical guarantee.

\subsection{Optimality Gap}

In this subsection, we will explain the detail of optimality gap $\epsilon$ of approximated $\epsilon$-best response in NFSP. \cite{FSP} has dicussed the optimality gap $\epsilon$ and left a corollary as corollary \ref{co:optimality gap}.
\begin{theorem} 
\label{co:optimality gap}
Let $\boldsymbol{\Gamma}$ be a two-player zero-sum extensive-form game with maximum payoff range $\overline{R}=\max_{\pi}{R^1\left(\pi\right)}-\min_{\pi}{R^1\left(\pi\right)}$. Consider a fictitious play proess in this game. Let $\Pi_t$ be the avearage strategy profile at iteration $t$, $B_{t+1}$ a profile of $\epsilon_t$-best responses to $\Pi_t$, and $\Pi_{t+1}=(1-\alpha_{t+1})\Pi_t+\alpha_{t+1}B_{t+1}$ the usual fictitious play update for some stepsize $\alpha_{t+1}\in(0,1)$. Then for each player $i$, $B_{t+1}^i$ is an $[(1-\alpha_{t+1})\epsilon_t+\alpha_{t+1}\overline{R}]$-best response to $\Pi_{t+1}$.
\end{theorem}
As the authors mentioned in \cite{FSP}, the value $[(1-\alpha_{t+1})\epsilon_t+\alpha_{t+1}\overline{R}]$ bounds the absolute amount by which reinforcement learning needs to improve the best response profile to achieve a monotonic decay of the optimality gap $\epsilon_t$. The authors relies $\epsilon_t$ decaying asymptotically on $\alpha_{t}\rightarrow0$.

The optimization of optimality gap $\epsilon$ is the focus of our work. We find that optimality gap could be reduced monotonically by utilizing regret matching method, thus let the optimality gap decay in a faster way than original NFSP, leading to better performance.

\subsection{Method and Proof}

Focusing on optimality gap $\epsilon$, we have found a new method that can make $\epsilon$ decay monotonically instead of asymptotically by replacing NFSP's best response computation with regret matching method. Optimality gap $\epsilon$ is formed in a way of regret $\omega$ and by making regret decaymonotonically, optimality gap reduces monotonically.

Regret $\omega=\{R^i(a,\pi^{-i})-R^i(\pi^i, \pi^{-i})\}_{a\in A}$ is the $m$-dimension regret vector of the algorithm at a specified iteration, where $A$ is the action profile on this information state and $m$ is the dimension of actions that could be taken. Regret has a similar definition with optimality gap $\epsilon$: $\epsilon\geq R^i(b^i_\epsilon(\pi^{-i}), \pi^{-i})-R^i(\pi^i, \pi^{-i})$, the overall regret of $\omega$ forms optimality gap $\epsilon$.

Therefore, minimizing the overall regret of $\omega$ could lead to the convergence of optimality gap.

Here we provide our proof of convergence of regret:

At first, we would like to provide two definitions about potential function and $P$-$regret$-$based$ $strategy$ as as definition \ref{potential function} and definition \ref{p regret base}. They are basis of our proof.

\begin{definition}
\label{potential function}
Given $D=\mathbb{R}^m$ the closed nagative orthant associated to the set of moves of player 1 and $\mathbb{R}$ the regret space. $P$ is a potential function for $D$ if it satisfies the following set of conditions:

(i) $P$ is a nonnegative function from $\mathbb{R}^m$ to $\mathbb{R}$

(ii) $P(\omega)=0$ iff $\omega\in D$

(iii) $\nabla P(\omega)\ge 0, \forall\omega\notin D$

(iv) $\left \langle\nabla P(\omega),\omega  \right \rangle > 0, \forall\omega\notin D$
\end{definition}
\begin{definition}
\label{p regret base}
$P$-$regret$-$based$ $strategy$ is defined by:

$\Pi_{t+1}-\Pi_{t}=\frac{1}{t+1}[\varphi_{t+1}(\omega)-\Pi_{t}]$ where

(i) $\varphi(\omega)=\frac{\nabla P(\omega)}{|\nabla P(\omega)|}\in X\ wherever \omega\notin D\ and\ \varphi(\omega)=X\ otherwise$, 
$X$ is the set of probabilities of actions, and $|\nabla P(\omega)|$ stands for the $L_1$ norm of $\nabla P(\omega)$. 
\end{definition}

Note that the update of $\varphi(\omega)$ is the same as regret matching method, and the $P$-$regret$-$based$ $strategy$ can also refer to the strategy update rule of NFSP if $\varphi(\omega)$ forms the best response.

Let $\omega^+$ be the vector with components $\omega^+_k=max(\omega_k,0)$. Define potential function $P(\omega)=\sum_k(\omega^+_k)^2$. Note that $\nabla P(\omega)=2\omega^+$, hence, $P$ satisfies the conditions (i)-(iv) of definition \ref{potential function}.

Under potential function $\nabla P(\omega)=2\omega^+$ and $\varphi(\omega)$, we provide theorem \ref{co:converge} of blackwell's framework. This theorem is summarized from \cite{Proof2}, which has been proved that if theorem \ref{co:converge} is satisfied, the strategy's regret will converge to 0.
\begin{theorem}
\label{co:converge}
Blackwell's framework makes the $P$-$regret$-$based$ $strategy$ approach the orthan $D$. In other words: If the potential function is defined as $P(\omega)=\sum_k(\omega^+_k)^2$, and the strategies of the players are independent, and the average strategy of a player statisfies $\Pi_{t+1}-\Pi_{t}=\frac{1}{t+1}[\varphi_{t+1}-\Pi_{t}]$, then the regret $\omega$ converges to zero.
\end{theorem}

Now that the update process of NFSP is $\Pi_{t+1}-\Pi_{t}=\alpha_{t+1}[B_{\epsilon_{t+1}}-\Pi_{t}]$ where $\alpha_{t+1}=\frac{1}{t+1}$, we replace best response $B$ with $\varphi$. The blackwell's framework is then satisfied, and as a result, regret $\omega$ is able to converge to zero.

Not like corollary \ref{co:optimality gap} relying optimality gap $\epsilon$'s decay on the decay of $\alpha$, our method applies regret matching to satisfy blackwell's framework (theorem \ref{co:converge}) to make regret and optimality gap converge to zero, which has a stronger guarantee than $\overline{R}$ and the decay of $\alpha$.

Involving regret matching method could lead our strategy update to a stronger guarantee to converge and reach smaller regret every iteration, which is not guaranteed by NFSP framework itself. Therefore, we find a possability to compute our best response using regret minimization based method.
\section{Introducing Regret Matching to NFSP}
This section is about the details of our method and algorithm. As is proved in section 3, our method is to replace best response computation with regret matching method. We applay Advantage-based Regret Minimization (ARM)~\cite{ARM}——a regret-matching-based reinforcement learning algorithm——as the regret matching method.

\begin{figure}[h]
  \centering
  \includegraphics[width=\linewidth]{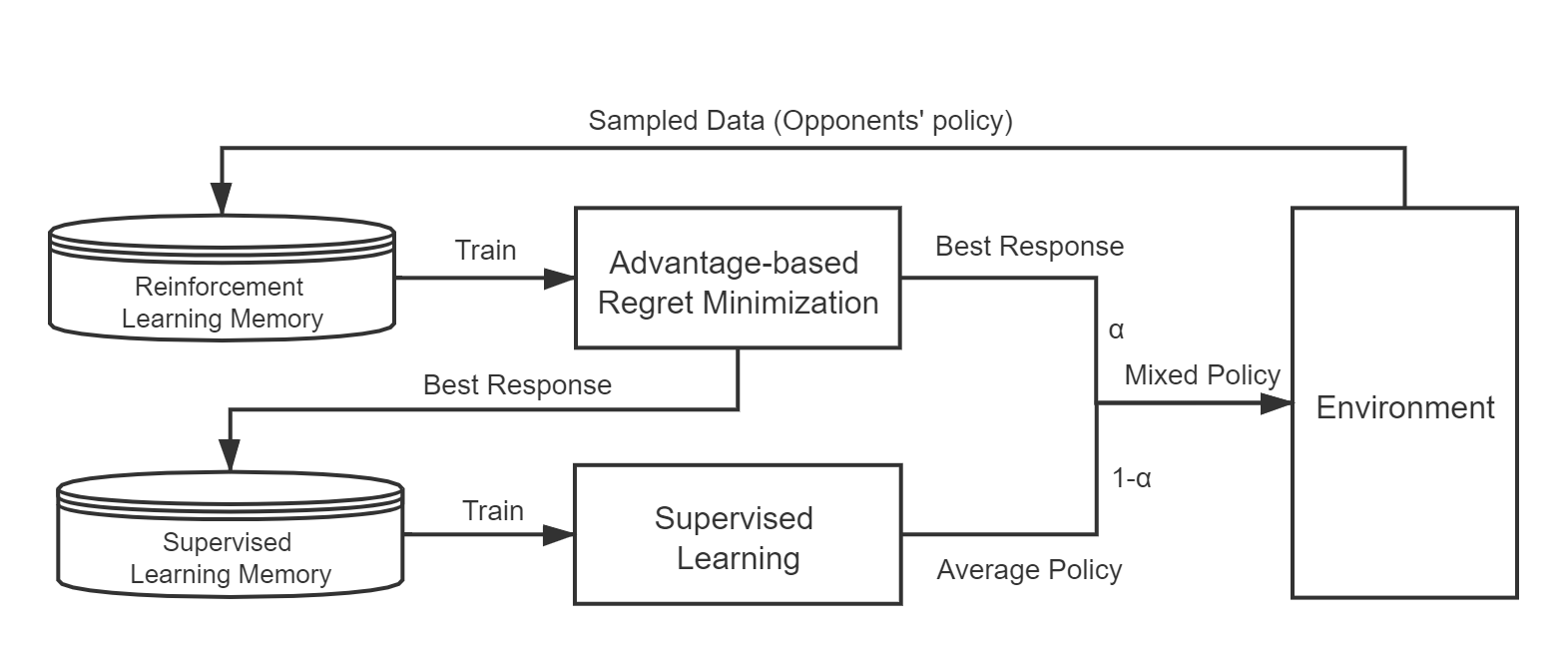}
  \caption{the overall structure of our method (one agent)}
  \label{fig:NFSP}
\end{figure}

It is simple to apply ARM to NFSP for ARM is a deep reinforcement learning method, similar to DQN in NFSP. In the fictitious self-play update process, we let ARM produce best response to the opponents' average strategy.

To make it clear, unlike NFSP, ARM follows a strict update rule of fictitious self-play. At iteration $k$, we sample from a mixture policy of $\left(1-\alpha\right)\Pi_k+\alpha B_{k+1}$, where $\Pi_k$ is the average policy at iteration $k$, $B_{k+1}$ is the best response to $\Pi_k$ (produced by ARM). Then ARM learns and optimizes the best policy $B_{k+2}$ to $\Pi_{k+1}=\left(1-\alpha\right)\Pi_k+\alpha B_{k+1}$, which is the average policy profile of next sampling iteration. Note that the $\alpha$ is a fixed number corresponding to the anticipatory parameter $\alpha$ of NFSP. We maintain numbers counting the recorded episodes and the update iterations. The number of episode refers to the hands of games, when a game is played and finished, the number of episode counts. The number of update iterations refers to the iteration of the algorithm's update. The number of update iteration counts when a particular number of timesteps are recorded, at the same time, the two networks will update to new best response and new average policy. After 10K episodes or an update iteration, the evaluation is executed.

As for ARM's update detail, we would like to refer the reader to ~\cite{ARM}.

Our algorithm is summarized in Algorithm 1, where RL refers to reinforcement learning and SL refers to supervised learning. 

\begin{algorithm}[tb]
\caption{Our method building ARM in NFSP} 
\begin{algorithmic}[1]
\State Initialize game and execute an agent via AGENT for each player in the game
\Function{AGENT}{}
\State Initialize RL replay memories $\mathcal{M}_{RL}$
\State Initialize SL replay memories $\mathcal{M}_{SL}$
\State Initialize ARM: $\theta_0, \omega_0\leftarrow arbitrary$ 
\State Initialize anticipatory parameter $\alpha$
\For{each episode}
\State policy $\pi\leftarrow
\begin{cases}
ARM(), &with probability\ \alpha\\
\Pi, &with probability\ 1-\alpha
\end{cases}$
\State Observe initial information state $s_1$ and reward $r_1$
\For{$t=1, T$}
\State Sample action $a_t$ from policy $\pi$
\State Execute action $a_t$ in game
\State Observe reward $r_{t+1}$, information state $s_{t+1}$
\State Store $(s_t, a_t, r_{t+1}, s_{t+1})$ in $\mathcal{M}_{RL}$
\If{agent follows best response policy}
\State Store $(s_t, a_t)$ in SL memory $\mathcal{M}_{SL}$
\EndIf
\State Update SL $\Pi$ with Memory $\mathcal{M}_{SL}$
\EndFor
\If{$\mathcal{M}_{RL}$ reaches a determined length}
\State execute ARM\_UPDATE to update ARM
\EndIf
\EndFor
\EndFunction

\Function{ARM}{observation o}
\State At update iteration $t$
\State $b_t(a|o)\leftarrow nomalize(Q^+_t(o,a;\omega_t)-V_t(o;\theta_t))$
\State \Return current $b_t(o)$
\EndFunction
\Function{ARM\_UPDATE}{}
\State At update iteration $t$
\State Update ARM's parameters as ~\cite{ARM}
\State iteration $t \leftarrow t+1$
\EndFunction 
\end{algorithmic} 
\end{algorithm}

\section{Experiments}
In this section, we will describe details of the experiments and make some analysis about the results.

We aim to evaluate our method on both perfect-information games and imperfect-information games, considering evaluating the applicability on these two different types of games. Since the optimality gap is produced in $\epsilon$-best response, the exploitability would somehow be a form of $(\epsilon_1 + \epsilon_2)/2$ where $(\epsilon_1$ is player1's optimality gap. Our method reachs smaller optimality gap at every update iteration, which means the exploitability will be lower than original NFSP.

We have conducted several experiments mainly in three multi-agent environments in OpenSpiel, published by Deepmind. OpenSpiel is a platform containing a huge amount of multi-agent games like Go, Leduc Poker, breakthrough, etc, where algorithms can play against each other. In addition to several evaluating methods for algorithms, the platform also provides a substantial number of typical algorithms from views of reinforcement learning and game theory including CFR, NFSP.

The three environments we chose are Leduc Poker, Liar's Dice and Tic Tac Toe. 
Introductions of these environments are along with the details of each experiment.

\subsection{Experiments on Leduc Poker}

We have examined exploitability and winning-rate of the different algorithms in Leduc Poker. The curves of exploitability show that our method converges faster and better than NFSP, and is more stable than ARM.

Leduc Poker is a small poker variant similar to kuhn poker~\cite{Kuhn}. With two betting rounds, a limit of two raises per round and 6 cards in the deck it is much smaller. In Leduc Poker, players usually have 3 actions to choose from, namely {fold, call, raise},and depending on context, calls and raises can be referred to as checks and bets respectively. The betting history thus can be represented as a tensor with 4 dimensions, namely {player, round, number of raises, action taken}. In Leduc Poker which contains 6 cards with 3 duplicates, 2 rounds, 0 to 2 raises per round and 2 actions, the length of the betting history vector is 30, considering the game will end if one player folds, the actions in information state is reduced from 3 to 2. 

\begin{figure}[h]
  \centering
  \includegraphics[scale=0.58]{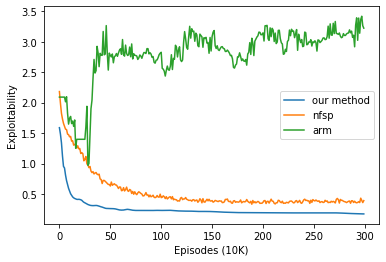}
  \caption{learning performance in leduc poker}
  \label{fig:ex_leduc}
\end{figure}

We trained the 3 algorithms separately to get their models and curves of exploitability, then let our trained algorithm play against trained NFSP to get the winning rate data to evaluate the performance of playing. While training, the hyper parameters of NFSP are set corresponding to the parameters mentioned in ~\cite{NFSP}, which are the best settings of NFSP in Leduc Hold'em.
The curves of exploitability while training is shown in Figure \ref{fig:ex_leduc}. As the curves show, to reach the same exploitability, our method costs much less time than original NFSP. Additionally, It is obvious that origin ARM’s strategy is highly exploitable, our work successfully transfers ARM from single agent imperfect-information games to zero-sum two-player imperfect information games.

As well as exploitability, the result of agents playing against each other is also examined to evaluate the performance of algorithms.
Therefore, in this part we let trained algorithms play against each other. To test the player position of the game whether an affact factor, we conduct 2 experiments. In each experiment, algorithms play for 40M hands of Leduc Poker and payoffs are recorded. Table \ref{tab:WinOfLeduc} shows the winning rate of player1 and loss rate, average payoffs as well. Note that in experiment 1, NFSP gets winning rate 46.7\%, and average payoff 0.01 (in zero-sum games), in experiment 2, our method gets winning rate 43.2\% and average payoff 0.15, it is obvious that our method is more likely to reduce loss when loses and maximize gains when wins.

\begin{table}[t]
  \caption{The results of algorithms' fight in Leduc Poker: our method's winning rate, loss rate and average payoff is shown}
  \label{tab:WinOfLeduc}
  \begin{tabular}{lllll}\toprule
    \textit{Player1} & \textit{Player2} & \textit{Winning Rate} & \textit{Loss Rate} & \textit{Average Payoff}\\ \midrule
    ours & NFSP & 39.6\% & 46.7\% & -0.01 \\
    NFSP & ours & 43.2\% & 42\% & 0.15 \\ \bottomrule
  \end{tabular}
\end{table}

\subsection{Experiments on Liar's Dice}
Experiment conduct on Liar's Dice is to evaluate if our method can outperform NFSP on other imperfect information environments.
Liar's Dice is a 1-die vesus 1-die variant of the popular game where players alternate bidding on the dice vlaues.

\begin{figure}[h]
  \centering
  \includegraphics[scale=0.58]{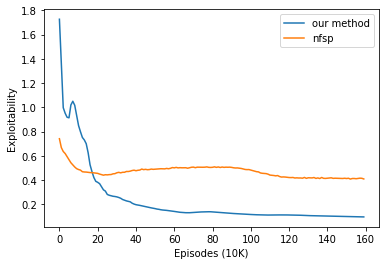}
  \caption{exploitability changes while training in liar's dice}
  \label{fig:ex_dice}
\end{figure}

The experiment design is similar to Leduc. Since the two turn-bases imperfect information game is similar while playing, we conduct this experiment following the settings in Leduc section.

The result of this experiment is also obvious.
The training exploitability is as figure \ref{fig:ex_dice} shows. Our method shows a greate advantage over NFSP, not only our method converges faster than the baseline, but also outperforms the baseline.

\subsection{Experiments on Tic Tac Toe}

We have examined exploitability and winning-rate of the different algorithms in Tic Tac Toe to see the algorithms' performance under perfect-information games, what's more, we have also examined the average reward playing against random agents every 1K episodes. In practice it presents an apparent difference between algorithms as this perfect-information board game environment is not that easy to be diturbed by random actions. 
The exploitability curves also show that our method converges faster and better than NFSP. In the meanwhile, our method outperforms NFSP at any player position of the two-player game.

Tic Tac Toe is a popular perfect-information board game. The game consists of a three by three grids and two different types of token, X and O. Each of the two players places his own token in order. The goal is to create a row of three tokens, which can be very tricky with both players blocking each other’s token to prevent each one from making a three-row of tokens. Once the three by three grids are full with tokens and nobody has created a row of three token, then the game is over and no one wins. In implementation, at the end of a game, an agent gets 1 if he wins and -1 for another agent, gets 0 if the game draws. 

Apparently, there exsits an advantage of playing in first place. While the first player intends to form the three-token-line all the time, the second player often has to defend from player1, therefore, for player2 the best strategy is to draw sometimes. That's why we consider draw rate when the algorithm plays at in disadvantageous position.

\begin{figure}[h]
  \centering
  \includegraphics[scale=0.58]{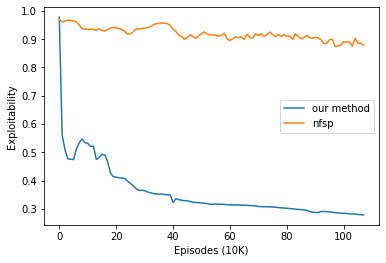}
  \caption{exploitability curves of training in tic tac toe}
  \label{fig:ex_tictac}
\end{figure}

We have trained our algorithm and NFSP in tic tac toe and the result is shown in Figure \ref{fig:ex_tictac}. It is clear that our methods outperforms NFSP in exploitability convergence.

\begin{figure}[h]
  \centering
  \includegraphics[scale=0.58]{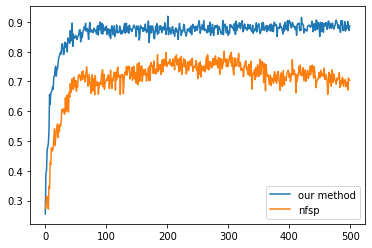}
  \caption{learning curve at player position 1}
  \label{fig:learn_p1}
\end{figure}

\begin{figure}[h]
  \centering
  \includegraphics[scale=0.58]{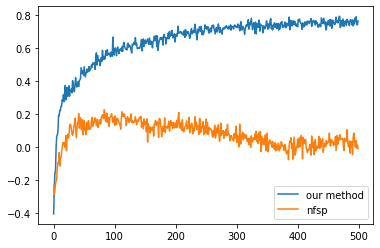}
  \caption{learning curve at player position 2}
  \label{fig:learn_p2}
\end{figure}

Since there exists an advantage of player who places his token first, we decided to examine the learning curves of the 2 conditions. A learning curve is a sequence of average rewards produced by playing against random agent at every 1K episodes. 
Figure \ref{fig:learn_p1} shows the learning curves with the first player position and Figure \ref{fig:learn_p2} shows the learning curves with the second player position. It is easy to see that our method performs better than NFSP in this type of game, including stability, convergence speed. 

After the algorithms are well trained, we let them fight against each other in different positions. Table \ref{tab:WinOfTTT} is the result of fighting part. Obviously, when our method plays at the player 1 position, the average reward could reach as high as 0.80 and the winning rate reaches 83.7\%. The more exciting thing is that even the first-player-advantage exists when our method plays at player 2 position, the average reward could still reach 0.658 and the winning rate is 67.4\% of wins and 30.9\% of draws. Note that considering the first-player-advantage, the goal for player2 is to win or to prevent player1 from winning. Therefore, a proportion of draw rate would be taken into account, which means the actual winning rate of player2 is more than pure winning rate. But even we use pure winning rate to evaluate the player 2 algorithm, our method still wins more than 50\% times.

\begin{table}[t]
  \caption{the results of algorithms' fight in tic\_tac\_toe}
  \label{tab:WinOfTTT}
  \begin{tabular}{llll}\toprule
    \textit{Player1} & \textit{Player2} & \textit{Winning Rate} & \textit{Loss Rate} \\ \midrule
    ours & NFSP & 83.7\% & 4.2\% \\
    NFSP & ours & 67.4\% & 2\%\\ \bottomrule
  \end{tabular}
\end{table}

\section{Conclusion}

In this paper, we focus on the optimality gap of NFSP and prove that this value could be guaranteed to decay monotonically by utilizing regret minimization methods.

Following the proof, we present our method by placing best response computate of NFSP with ARM. In our method, ARM applies regret matching to calculating best response to the opponents' strategy. As a result of this combination, the convergence speed is fastened and in practice the performance of our method is also better than original NFSP implemented in OpenSpiel.

Our work is an attempt to combine the two mainstream solutions Self-Play and CFR to approach approximate Nash equilibrium. It is proved that the performance of self-play methods could be optimized by utilizing a regret matching methods. That is, the application of regret minimization method to FSP framework algorithms is effective and could certainly increase the performance.

There is still a lot remained to explore. 
Our further study is to evaluate more algorithms combining FSP and regret minimization reinforcement learning methods.
Moreover, ARM applies to image-input games like Minecraft, Doom mentioned and we could also try to extend our methods to image-input environments, thus make it suitable for a larger range of games.

\bibliographystyle{named}
\bibliography{nfsp_arm_paper}

\end{document}